\documentclass[letterpaper, 10 pt, conference]{ieeeconf}  % Comment this line out
\IEEEoverridecommandlockouts                              % This command is only
\usepackage{graphics} % for pdf, bitmapped graphics files
\usepackage{amsmath} % assumes amsmath package installed
\usepackage{bm}
\usepackage{amsfonts}
\usepackage{textcomp}
\usepackage{xcolor}

\usepackage{multirow}
\usepackage[pdftex]{graphicx}
\usepackage[caption=false,font=footnotesize,labelfont=sf,textfont=sf]{subfig}
\title{\LARGE \bf
A Network-based Multimodal Data Fusion Approach for Characterizing Dynamic Multimodal Physiological Patterns
}

\author{Miaolin Fan$^{1}$, Chun-An Chou$^{1}$, Sheng-Che Yen$^{2}$ and Yingzi Lin$^{1}$% <-this % stops a space
\thanks{*This work was supported by the Northeastern TIER 1 Seed Grant Program.}% <-this % stops a space
\thanks{$^{1}$Department of Mechanical and Industrial Engineering,
Northeastern University, Boston, MA, 02115, USA. Corresponding author contact
        {\tt\small ch.chou@northeastern.edu}}%
\thanks{$^{2}$Departments of Physical Therapy, Movement \& Rehabilitation Science, Northeastern University, Boston, MA 02115, USA
        }%
}

\begin{document}

\maketitle
\thispagestyle{empty}
\pagestyle{empty}

%%%%%%%%%%%%%%%%%%%%%%%%%%%%%%%%%%%%%%%%%%%%%%%%%%%%%%%%%%%%%%%%%%%%%%%%%%%%%%%%
\begin{abstract}

Characterizing the dynamic interactive patterns of complex systems helps gain in-depth understanding of how components interrelate with each other while performing certain functions as a whole. In this study, we present a novel multimodal data fusion approach to construct a complex network, which models the interactions of biological subsystems in the human body under emotional states through physiological responses. Joint recurrence plot and temporal network metrics are employed to integrate the multimodal information at the signal level. A benchmark public dataset of is used for evaluating our model.

\end{abstract}

%%%%%%%%%%%%%%%%%%%%%%%%%%%%%%%%%%%%%%%%%%%%%%%%%%%%%%%%%%%%%%%%%%%%%%%%%%%%%%%%
\section{Introduction}
Daily activities of human body are performed through the joint functioning of biological subsystems, including nervous, muscular, respiratory, etc. Extensive attention has been devoted into developing methods for utilizing the rich information collected from human body via multiple sources, while each source of information is referred to as a modality. The major challenge of \textit{multimodal data fusion} of human physiological activities is to characterize the nonlinear, time-varying interrelationship among subsystems. To address this challenge, a trending approach in recent years is to construct a mapping from multivariate time series to complex networks for modeling the dynamic patterns of interactions among components in a complex system \cite{Gao_2017,Amigo_2018}. In network models, the nodes represent the components of system and the edges represent the (dis)similarity or dependency relationships among any pairs of components (e.g., Euclidean distance, mutual information, causality, etc.). On the other side, the topological structures of network models can be described through a variety of quantitative metrics, which allows hybrid pattern recognition methods to classify the mental/physical states of the human body through physiological activities.
\\
Despite the strong potential of network approaches for analyzing human body as a complex system, limitations still exists since the conventional distance measures are not perfectly suitable for capturing the (dis)similarity between the coupling dynamics for physiological time series \cite{Spiegel_2014}. Distance measures for quantifying the interrelationships among multivariate time series were introduced in previous study (for a comprehensive and most recent review, please refer to \cite{Gao_2017b}). In particular, considering the nonlinear and non-stationary properties of human physiological signals, we selected recurrence plot (RP) \cite{Eckmann_1987} to characterize the pair-wise (dis)similarity based on the chaotic behaviors of physiological time series. Through the mapping from time series to RP, quantification metrics (recurrence quantification analysis, RQA \cite{Marwan_2002}) are available to describe the chaotic features of a complex system, e.g. periodicity and predictability (determinism). Moreover, Romano \cite{Romano_2007} extended RQA from univariate to multivariate time series. Feldhoff et al. developed a network-based method to identify the inter-system coupling relationships \cite{Feldhoff_2012}. These previous studies have enabled (1) a new class of quantitative distance measurements for nonlinear time series analysis based on their synchronization attributes; (2) the integration of network-based analysis and multivariate pattern recognition to time series data. However, as previous studies used homogeneous temporal data collected from systems/subsystems driven by similar sources, the capability of this type of methods has not been fully investigated on complex systems consisting of multivariate time series with heterogeneous response patterns, temporal scales and resolutions. Most existing multimodal data fusion methods, to the best of our knowledge, integrate the information from multimodal physiological data based on \textit{feature} or \textit{decision} level.

In this study, we present a recurrence-plot based method to fuse information from multivariate time series on the \textit{signal} level in order to recover the nonlinear coupling relationships among biological subsystems while preserve the heterogeneous temporal scales in physiological responses of different subsystems. Moreover, we intend to examine whether if our method is capable to characterize interrelation patterns of the human body under different emotional states using the physiological patterns represented by time-varying networks.
% \textcolor{red}{The remainder of this paper is organized as follows. The overall idea is described in Section \ref{method} using toy-example. We test our approach on the emotion recognition problem using multimodal physiological signals collected from human subjects during music video watching tasks and presented the experimental results in Section \ref{exp}. Finally, we summarize the significance and limitations of the existing method as well as proposing potential directions for future studies in Section \ref{conclusion}.} 
\section{Method}
\label{method}
\begin{figure*}[ht!]
\centering
\includegraphics[width=\textwidth]{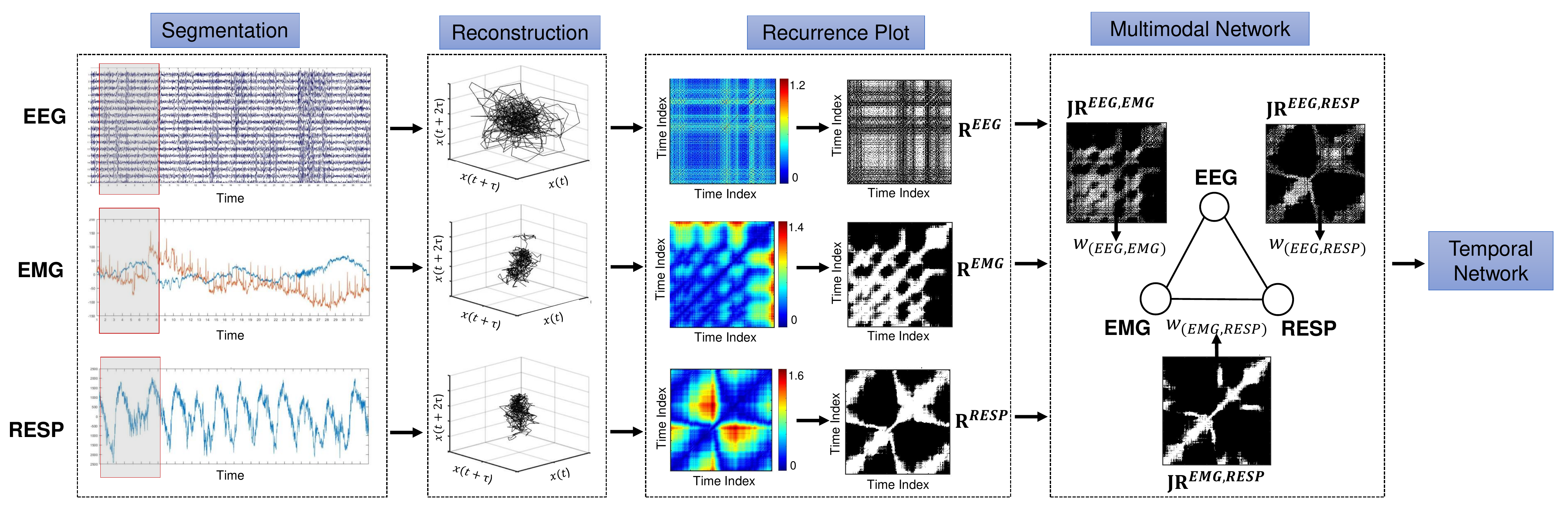}
\caption{A step-by-step demonstration of our proposed multimodal data fusion framework using sample data from three modalities (EEG, EMG and Respiration) in DEAP dataset \cite{Koelstra_2012}. The segmented signals are reconstructed in the state space and mapped into RPs. JRPs are obtained by multiplication of each pair of RPs from all modalities, in which JDET and JLAM account for the overall percentage of points on diagonal and vertical lines. Assigning the JDET/JLAM as the weight to each edge between the corresponding pair of modalities creates a static network for each time window, and concatenating the networks of all time windows constructs a temporal network to represent the entire trial.}
\label{fig:flow}
\end{figure*}
In this section, we demonstrate a network model for mapping multimodal time series to complex dynamic networks, where the nodes in the network represent modalities and edges are defined based on the directional coupling between modalities. Fig. \ref{fig:flow} illustrates the methodological flow. The network construction includes the following steps: (1) the reconstruction from time series to time-varying graphs using time-delayed embedding theorem \cite{Takens_1981}; (2) the RP construction for each channel \cite{Eckmann_1987}; (3) the extension of RP to multivariate time series using joint RP (JRP) \cite{Romano_2004}; and (4) the extraction of time-varying graph theoretic network metrics for classifying different emotional states.

In order to characterize the nonlinear coupling strength among multivariate time series, Determinism (DET) and Laminarity (LAM) are extracted from JRP \cite{Romano_2004}. For a univariate time series signal, the patterns of recurrences are depicted by RP \cite{Eckmann_1987}, which is a binary matrix representing the neighborhood relationships among all state vectors during a certain period of time. Sequentially, the information fusion between any pairs of channels/variables is conducted by multiplying two RPs point-to-point, generating another binary matrix (JRP). JRP can be converted into a network by treating the binary matrix as the adjacency matrix $A$ for a network $G$, as shown in Fig. \ref{fig:flow}. $G$ shows the temporal dependency pattern of a system using graph representations. The mathematical details are explained in the following sections.

\subsection{From Univariate to Multivariate Nonlinear Time Series Analysis}
RP is a visualization tool to represent the temporal dependency relationships between all states in a time series data using a binary, squared matrix \cite{Eckmann_1987}. Suppose the state of a system (or modality) $X$ at time \textit{i} and {j} is represented by $\textbf{x}_i, \textbf{x}_j$, the recurrences can be recorded by the following binary function:
\begin{equation}
\textbf{R}_{i,j}^{X} = \Theta (\epsilon_X -{||\textbf{x}_i - \textbf{x}_j||_1}), \textbf{x}_i \in \mathbb{R}^m, \textit{i}, \textit{j} = 1,...N,
\end{equation}
where $\Theta$ is a Heaviside function and a black dot is assigned to coordinates $(i,j)$ on RP if ${R}_{i,j}^{X} = 1$, indicating at time $i$ and $j$ the state of the system is sufficiently close (within the system threshold $\epsilon_X$). 
\\
Although the original method was developed for a single time series, variations of RP included consideration of multivariate time series from different aspects. As the goal is to best capture the dynamic coupling between multiple modalities, JRP is employed which represents when a recurrence occurs simultaneously in two or more time series \cite{Romano_2004}. For any two systems $X$ and $Y$, the JRP is obtained by the equation:
\begin{equation}
\textbf{JR}_{i,j}^{X,Y} = \Theta (\epsilon_X -{||\textbf{x}_i - \textbf{x}_j||}) ~ \Theta (\epsilon_Y -{||\textbf{y}_i - \textbf{y}_j||}).
\end{equation}

Moreover, as JRP characterizes the \textit{simultaneous occurrence} of recurrences \cite{Marwan_2002}, the joint recurrence quantification analysis (JRQA) defines a set of metrics based on JRP for quantifying nonlinear dynamics underlying bivariate coupling time series. In the present study, two selected metrics are computed based on JRP, which are defined as below \cite{Romano_2004}:
\subsubsection{Determinism (DET)}
describes the predictability of systems. DET of $\textbf{R}_{i,j}^{X}$ is defined as the average length of diagonal lines with minimal length $l_{min}$. Similarly, JDET can be computed to present how predictable the coupling behaviors are for coupling systems X, Y based on JRP $\textbf{JR}_{i,j}^{X,Y}$. In the present study, we select the $l_{min} = 3$, which means a diagonal line need to have at least three points in a row.
\subsubsection{Laminarity (LAM)}
Laminarity describes the average time of staying in a consistent state. This is an analog of DET but defined based on vertical lines. LAM is computed as the average length of vertical lines with minimal length $v_{min}$, and comparably, JLAM is computed for $\textbf{JR}_{i,j}^{X,Y}$. In the present study, we define the $v_{min} = 3$.

\subsection{Constructing Time-varying Graphs and Identifying Coupling Relationships}
\label{Sec2b}
In order to capture the time-varying dynamics and perform a signal-level data fusion, a sliding window is applied to each channel throughout the time series as the first step. The segmented time series data is used for reconstructing phase space trajectory and constructing RP, which describes the recurrence patterns of each modality. Then, the inter-modality connections are built based on the simultaneous recurrences of two modalities by constructing JRP using pair-wise multiplications of RPs. Finally, JRQA metrics are computed in order to quantify the coupling of periodic/deterministic behaviors for any two subsystems (modalities) invariant to the original temporal scales of time series. This desirable feature enables the signal-level data fusion regardless of varying temporal scales. The resulting JRQA metrics are converted into edge weights for the corresponding pair of modalities. Note that each node can either present a channel or a modality; for the latter case, if one modality has multiple channels, the nodes of all channels are merged as one node to present the modality by taking the avereage weights of each inter- or intra-system edge. In this study, channels from the same modality are merged, such that each node represents a modality. Lastly, after constructing the network representation $G^T$ for each time window $T = \{x_1,x_2,...\}$, a time-varying network is created by concatenating the static networks obtained by applying a sliding window throughout an $m$-dimensional time series $\{x_1,x_2,...x_t,...\}$, where each $x_t$ is a column vector of $m$ components.

\subsection{Time-varying Graph Theoretic Metrics for Multimodal Pattern Characterization}
\label{Sec2c}
Temporal networks, or time-varying networks, are used in our model for characterizing the time-ordered sequential changes among physiological subsystems \cite{Nicosia_2013}. Some metrics are defined as the generalization of the static network metrics, while the temporal changes are considered as an additional dimension. The selected time-varying network metrics are implemented in MATLAB 2017b using a toolbox \cite{Sizemore_2017}. The interpretation of selected graph theoretic metrics is as follows. 
\subsubsection{Network efficiency}
defines how efficiently information could spread throughout the network in temporal scale. It is a global measure for the entire network.
\subsubsection{Time-varying shortest path}
defines how fast can two nodes be reachable for each other in temporal scale. The unit for steps are time intervals.
\subsubsection{Number of fastest paths}
defines how many fastest time-varying shortest paths exist in between two nodes.
\subsubsection{Temporal small-worldness and temporal correlation coefficients}
generalizes of small-worldness and clustering correlation coefficients to time-varying network.
\subsubsection{Connectedness}
is a measure of reachability. Strong-connected nodes refer to two nodes which are reachable from both forward and backward temporal directions (i.e. paths from node \textit{i} to \textit{j} and \textit{j} to \textit{i} are both available in time order). Weak-connected nodes refer to two nodes which are reachable from only one temporal direction.
\\

\section{Experimental Results}
\label{exp}
\subsection{Dataset Description and Pre-processing}
\label{dataset}
Our model was evaluated using five selected subjects from DEAP, a benchmark multimodal database in which EEG recordings and peripheral physiological signals were collected and published by Koelstra and colleagues \cite{Koelstra_2012}. A 10-20 International system of 32 AgCl electrodes was used for EEG data recording at a sampling rate of 512 Hz. Multimodal physiological responses used for our study include Galvanic skin response (GSR) measured form left middle and ring finger, Zygomaticus and Trapezius Electromyography (zEMG and tEMG), Electrooculogram (EOG) and respiration (RESP).

In the experiment phase, each subject was instructed to watch music videos while their physiological responses were monitored by wearable sensors. Self-Assessment Manikin (SAM) \cite{Bradley_1994} questionnaires were distributed to collect self-reported emotional scores (0 -- 9) regarding these videos based on four dimensions: arousal, valence, liking, and dominance. Then, a three-class classification problem was formulated by discretizing continuous scores of arousal and valence into low (1 -- 4), medium (4 -- 6) and high (6 -- 9). The pre-processed signals (as described in \cite{Koelstra_2012}) have been down-sampled to 128 Hz and segmented into 40 trials of one-minute long. In addition, EMG signals were high-pass filtered at 10 Hz, and all other signals were bandpass filtered at 4 -- 40 Hz. The sliding window for segmentation was five seconds with 20\% overlap.
\\

\subsection{Feature Extraction}
Constructing JRP from bivariate time series data requires the selection of time-embedding parameters, which is conducted as suggested in previous literature \cite{Marwan_2002}. Furthermore, since we intend to preserve as much information as possible from each channel, the JRP-based metrics are computed between each pair of channels to preserve interactions both across and within each modality. For example, the edge weight between EEG and EMG is computed by taking the overall average for all existing edges between EEG and EMG channels; the within-modality interaction is estimated by averaging the edge weights within EEG and EMG respectively. Lastly, the time-varying graph metrics, as described in Section \ref{Sec2c}, are computed as features and fed to selected machine learning classifiers for training classification models.
\subsection{Emotional States Classification and Validation}
The selected subjects have at least 8 trials in each class (low, medium, high) of emotional states. Since many features are expected to be redundant or irrelevant to the prediction target, $L1$-Norm regularized logistic regression model (LASSO, \cite{Tibshirani_1996}) is used to simultaneously perform feature selection with model fitting by forcing small coefficients to be zeros. Two models are trained and evaluated independently for arousal and valence.

Classification performances are summarized in Table \ref{tb_performance}. The best model for classifying valence uses JDET as edge weights for the graphs and achieves 52.50\% for three-class classification, while the best model for classifying arousal achieves 57.9\% using JLAM. Five-fold cross-validation is used for evaluating the generalizability on unobserved data. Our algorithm is implemented in MATLAB 2017b using customized code, while third-party toolboxes are used for computing features and training/testing classification models \cite{Soleymani_2017,Sizemore_2017,Qian_2013}.

\begin{table}[!t]
\caption{Comparison of Performances between networks Constructed based on JDET and JLAM}
\label{tb_performance}
\centering
\begin{IEEEeqnarraybox}[\IEEEeqnarraystrutmode\IEEEeqnarraystrutsizeadd{2pt}{1pt}][b][\columnwidth]{s+s+t+t}
\IEEEeqnarraydblrulerowcut\\
\IEEEeqnarrayseprow[2pt]{}\\
Study & Machine Learning & Valence & Arousal \IEEEeqnarraystrutsizeadd{0pt}{-1pt}\\
\IEEEeqnarrayseprow[2pt]{}\\
 & Model & Accuracy & Accuracy \IEEEeqnarraystrutsizeadd{0pt}{-1pt}\\
\IEEEeqnarrayseprow[2pt]{}\\
\IEEEeqnarrayrulerow\\
\IEEEeqnarrayseprow[2pt]{}\\
Chung and Yoon \cite{Chung_2012} & Bayesian & 53.40\% & 51.00\% \\
Tripathi \cite{Tripathi_2017} & CNN & 66.79\% & 57.58\%\\
Our model & LASSO & 52.50\% & 57.90\%\\
\IEEEeqnarrayseprow[4pt]{}\\
\IEEEeqnarraydblrulerowcut
\end{IEEEeqnarraybox}
\end{table}

\subsection{Discussions}
\begin{figure}[!t]
\centering
\subfloat[]{\includegraphics[width=0.5\textwidth]{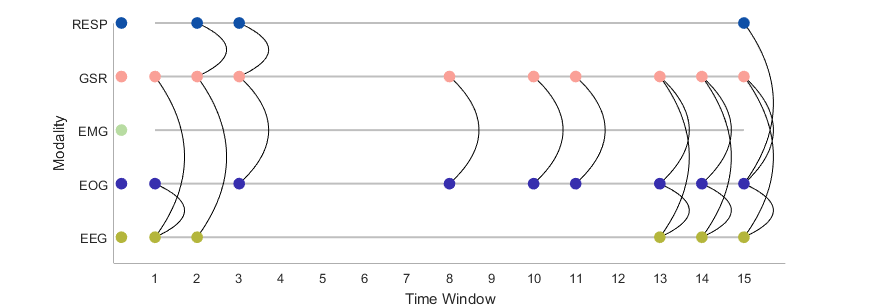}%
\label{fig2a}}
% \hfil
\hskip -1pt plus 1fil
\subfloat[]{\includegraphics[width=0.5\textwidth]{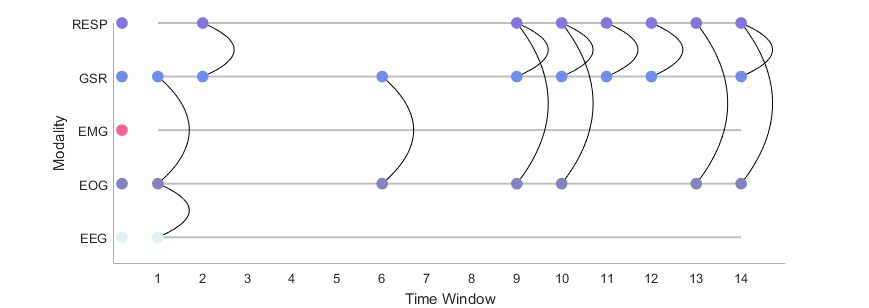}%
\label{fig2b}}
% \hfil
\hskip -1pt plus 1fil
\subfloat[]{\includegraphics[width=0.5\textwidth]{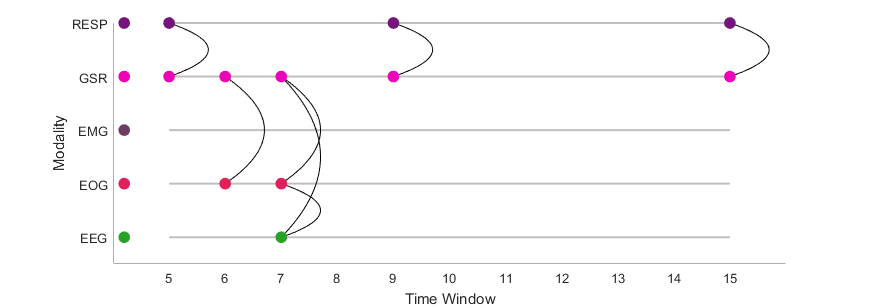}%
\label{fig2c}}
% \hfil
\hskip -1pt plus 1fil
\subfloat[]{\includegraphics[width=0.5\textwidth]{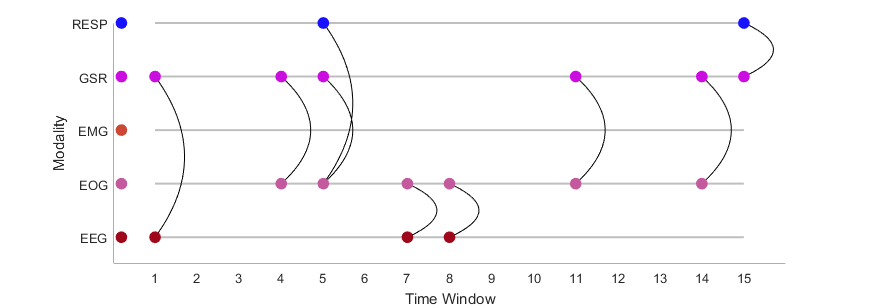}%
\label{fig2d}}
\caption{A visualization of temporal networks in each emotional state using examplar data from s02. Each node represents one modality (e.g. EEG, EMG, etc.), and each edge represents an observed synchronization at present time window. These plots show the dynamic connectivity patterns change over time (along the horizontal axis) under each type of emotional state, including (a) High valence, high arousal. (b) Low valence, low arousal. (c) High valence, low arousal. (d) Low valence, high arousal.}
% or (a), (b) can each carry a complete description:
%\caption{(a) First description. (b) Second description.}
\label{fig2}
\end{figure}

As an example, the average patterns of dynamic synchronization among modalities for one subject (s02) are shown in Fig. \ref{fig2}. Note that each individual may have different connectivity patterns for emotional states. An interesting observation is that EMG has rarely been connected to any other modalities during all conditions. GSR and EOG appear to have more connections in high arousal scenarios, while respiration and ECG have more connections in low arousal scenarios. These patterns may reflect the physiological responses associated with psychological excitement or calm.

Compared to two previous studies \cite{Chung_2012,Tripathi_2017} for three-class (high, medium, low) classification tasks on the same dataset, our model is comparable on classifying the emotional states in terms of the classification accuracy while considering the random selection level as 33.33\% (as shown in Table \ref{tb_performance}). While previous studies used more conventional and black-box models, the extracted network features are useful to recognize the underlying system states and our model based on a linear LASSO classifier allows for intuitive insights into physiological responses of human affective states from a dynamic network perspective.

It is noteworthy that the proposed model produced more misclassified cases in the \textit{medium} level than in the \textit{low} and \textit{high} levels. Given that the emotional states are labeled based on self-report SAM scores, a potential source of variation is the subjective scoring of human subjects especially for the \textit{neutral} states. It is extremely difficult to train a model on a small sample when some cases located on the boundary of two classes are very close. Another limitation of our method is the connectedness of time-varying graphs. The computation of some graph theoretic metrics would become problematic when the graph is disconnected; therefore, the threshold selection is critical when converting the distance metrics to binary metrics.
%
% Also note that IEEE Sensors Letters uses the naming convention of
% fig1.eps, fig2a.pdf, etc., for submitted graphics files.
%
% \begin{figure}[h]
% \begin{subfigure}{0.5\textwidth}
% \includegraphics[width=0.9\linewidth]{highVhighA.png} 
% \caption{High valence and high arousal}
% \label{fig:subim1}
% \end{subfigure}
% \begin{subfigure}{0.5\textwidth}
% \includegraphics[width=0.9\linewidth]{lowVlowA.png}
% \caption{Low valence and low arousal}
% \label{fig:subim2}
% \end{subfigure}
% \caption{Caption for this figure with two images}
% \label{fig:image2}
% \end{figure}

% An example of a figure with subfigures using the subfig.sty package.
% (The subfig.sty package must be loaded for this to work.)
% The subfigure \label commands are set within each subfloat command,
% and the \label for the overall figure must come after \caption.
% \hfil is used as a separator to get equal spacing.
% Watch out that the combined width of all the subfigures on a 
% line do not exceed the column width or a line break will occur.
%

\section{Conclusion}
\label{conclusion}
In this study, we demonstrated a novel method for building time-varying networks as representations for time series signals. After extracting temporal graph-theoretic features based on both spatial and temporal coupling dynamics, the classification model was applied to recognize the emotional states of the complex human body system consisting of multimodal physiological time series. Our future work would extend the current model to a general approach which uses time-varying networks to characterize complex systems with various types of coupling relationships, allowing applications to different domains. Besides, the efficiency of information fusion in multiresolution, multimodal complex systems by approximate computations in network construction and metrics computation will be investigated more specifically.

%%%%%%%%%%%%%%%%%%%%%%%%%%%%%%%%%%%%%%%%%%%%%%%%%%%%%%%%%%%%%%%%%%%%%%%%%%%%%%%%

%%%%%%%%%%%%%%%%%%%%%%%%%%%%%%%%%%%%%%%%%%%%%%%%%%%%%%%%%%%%%%%%%%%%%%%%%%%%%%%%



% \section*{ACKNOWLEDGMENT}

% The preferred spelling of the word ÒacknowledgmentÓ in America is without an ÒeÓ after the ÒgÓ. Avoid the stilted expression, ÒOne of us (R. B. G.) thanks . . .Ó  Instead, try ÒR. B. G. thanksÓ. Put sponsor acknowledgments in the unnumbered footnote on the first page.

%%%%%%%%%%%%%%%%%%%%%%%%%%%%%%%%%%%%%%%%%%%%%%%%%%%%%%%%%%%%%%%%%%%%%%%%%%%%%%%%

% References are important to the reader; therefore, each citation must be complete and correct. If at all possible, references should be commonly available publications.

% \bibliographystyle{IEEEtran}
% \bibliography{main.bib}


\end{document}